\documentclass{article}

\usepackage{microtype}
\usepackage{graphicx}
\usepackage{booktabs} 

\usepackage{hyperref}

\usepackage[accepted]{icml2023}

\usepackage{amsmath}
\usepackage{amssymb}
\usepackage{mathtools}
\usepackage{amsthm}
\usepackage{tikz}
\usepackage{ulem}
\usepackage[capitalize,noabbrev]{cleveref}

\def\checkmark{\tikz\fill[scale=0.4](0,.35) -- (.25,0) -- (1,.7) -- (.25,.15) -- cycle;}

\newcommand{\addSylv}[1]{\textcolor{black}{#1}}
\newcommand{\addmoi}[1]{\textcolor{black}{#1}}
\newcommand{\addSido}[1]{\textcolor{black}{#1}}

\newcommand{\NFA}{\textsc{NFA}}
\newcommand{\Sigm}{\textsc{Sigm}_{\alpha}}
\theoremstyle{plain}

\theoremstyle{definition}

\theoremstyle{remark}

\usepackage[textsize=tiny]{todonotes}

\begin{document}

\twocolumn[
\icmltitle{Deep-NFA: a Deep \textit{a contrario} Framework for Small Object Detection}



\icmlsetsymbol{equal}{*}

\begin{icmlauthorlist}
\icmlauthor{Alina Ciocarlan}{yyy,comp}
\icmlauthor{Sylvie Le Hegarat-Mascle}{comp}
\icmlauthor{Sidonie Lefebvre}{yyy}
\icmlauthor{Arnaud Woiselle}{sch}
\end{icmlauthorlist}

\icmlaffiliation{yyy}{DOTA, ONERA, Paris-Saclay University, F-91123 Palaiseau, France}
\icmlaffiliation{comp}{SATIE, Paris-Saclay University, 91405 Orsay, France}
\icmlaffiliation{sch}{Safran Electronics $\&$ Defense, F-91344 Massy, France}

\icmlcorrespondingauthor{Alina Ciocarlan}{alina.ciocarlan@onera.fr}

\icmlkeywords{small object detection, deep learning, segmentation, a contrario, NFA, interpretability}

\vskip 0.3in
]



\printAffiliationsAndNotice{}  

\begin{abstract}
The detection of small objects is a challenging task in computer vision. Conventional object detection methods have difficulty in finding the balance between high detection and low false alarm \addmoi{rates}. \addmoi{In the literature, some methods have addressed this issue by enhancing \addSylv{the} feature map responses, but without guaranteeing robustness with respect to the number of false alarms induced by 
background elements}. To tackle this problem, we introduce an \textit{a contrario} decision criterion into the learning process to take into account the unexpectedness of small objects. This statistic criterion enhances \addSylv{the} feature map responses while controlling the number of false alarms (NFA) and can be integrated into any semantic segmentation neural network. Our add-on NFA module not only allows us to obtain competitive results for small target and crack detection tasks respectively, but also leads to more robust and interpretable results.

\end{abstract}

\section{Introduction}
\label{intro}
The detection of small objects, defined as having \addSylv{at least} one small ($1-5$ pixels) dimension, is a great challenge in computer vision.
This issue is of a key interest in many real-world applications, \addSylv{e.g.} in the defense and security fields for surveillance or in the medical field for early and accurate diagnosis. 
The rise of deep learning methods has led to impressive progress in object detection in the past decades, mostly thanks to their ability to extract non-linear features well adapted to the downstream task  \cite{ren2015faster,redmon2016you}. However, most state-of-the-art (SOTA) object detection methods perform poorly on small objects. \addmoi{Indeed, neural networks (NN) based on bounding box regression either have anchors too large for detecting small objects, or require a significant decrease in the IoU threshold, which leads to \addmoi{multiple detections of a same target~\cite{dai2021attentional}.}
Semantic segmentation methods are then preferred, although their performance remains limited.} This is first\addSylv{ly} due to the nature of the objects: their surface area is made of only few pixels, and they do not present a specific structure. Secondly, small objects are often  \addSido{partially hidden in complex and highly textured backgrounds}, leading to many false alarms. 
Moreover, dealing with small object detection results in learning from highly class-imbalanced datasets. Thus, in \addSido{a} semantic segmentation scheme, their features cannot be learned \addSylv{easily} \addSylv{due to the small number of samples} in comparison with the background class. 

Few approaches have been proposed to improve small object detection. Some of them focus on augmenting or oversampling the dataset~\cite{kisantal2019augmentation, akyon2022sahi}, while others focus on improving small object feature-enhancement. \addSylv{Among} the latter, we can cite Feature Pyramid Networks (FPN) and their variants~\cite{lin2017feature,deng2021extended}, \addSylv{whose} multi-scale approach is beneficial for the detection of objects of various sizes. Other works include attention mechanisms to learn long-range dependencies~\cite{wu2022uiu}, or use super-resolution to enhance feature responses of small objects~\cite{deng2021extended}. However, these methods do not take advantage of the unexpectedness of small objects with respect to the background, as one could do in an anomaly detection approach with, for example, one\addSylv{-}class classifiers~\cite{pmlr-v80-ruff18a}. This criterion is important to reduce the number of false alarms and thus find a better balance between precision and detection rate.

We \addmoi{therefore} propose a module, called NFA module, which takes into account the unexpectedness of small objects and enhances their feature responses by rejecting the background hypothesis.
Our module relies on \textit{a contrario} \addSylv{decision} methods, introduced by \addSylv{\cite{desolneux2003grouping}}. These methods allow to automatically \addSylv{derive a decision criterion} with regards to hypothesis testing and \addSylv{draw inspiration} from theories of perception, in particular the Gestalt theory~\cite{desolneux2007gestalt}. 
They are based on the Helmholtz principle, which states that a large deviation from a random pattern is probably due to the presence of a structure. 
Concretely, \addSylv{the} \textit{a contrario} detection methods consist in rejecting a naive model describing the background by \addSylv{minimizing} the Number of False Alarms (NFA, defined in Section~\ref{generalite_NFA}). One of the benefits of this \addSylv{kind of approaches} is that it does not rely on any prior knowledge about the image \addSylv{structures: it only assumes a naive model for the noise.} \addmoi{Moreover}, it allows \addSylv{for} the control of the NFA. 

Many \textit{a contrario} formulations have been proposed in the literature, relying on different naive models. \addSylv{An important distinction is the kind of considered images, namely either binary or gray-level images. In the first case, a widely used naive model is the uniform spatial distribution of the ``true'' pixels in the image lattice, leading to binomial distribution for the number of ``true'' pixels falling within any given parametric shape~\cite{desolneux2003grouping, HegaratMascle2019}. In the second case, a widely used naive model is the Gaussian distribution of the pixel gray-level values, leading to \addSido{chi-square} distribution for the sum of the square\addmoi{d} errors~\cite{Robin2010, ipol.2019.263}.} 
Our work will be based on\addSylv{~\cite{ipol.2019.263}}. 

In the literature, the \textit{a contrario} test is applied either on natural images directly or after extracting features from the image by classical image processing methods. 
This filtering step can easily be replaced by \addSylv{NN}. Indeed, when \addmoi{looking at feature maps of a NN} trained for detecting objects, \addSylv{the} objects to detect stand out against a background made of noise. It is possible to apply \textit{a contrario} detection on these feature maps, as done in\addSylv{~\cite{ipol.2019.263}}, \addmoi{but such a post-processing step appears suboptimal since it does not guide the training step.}
We therefore propose to include \addSylv{the chosen} \textit{a contrario} criterion in the training, through our NFA module. This module can be \addmoi{integrated into}
any segmentation NN, and can even take advantage of multi-scale information if the backbone allows it. 
Our main contributions can be summarized as follows:
\begin{enumerate}
    \item We propose a new module specifically designed for small object detection that takes into account the unexpectedness of an object thanks to an \textit{a contrario} decision criterion.
    \item We demonstrate the effectiveness of our method for infrared small target detection, which achieves state-of-the-art performance while also leading to more interpretable results.
    \item We extend our experiments to other backbone and application, namely crack detection, in order to show the generalization ability of the proposed method. 
\end{enumerate}

\section{Theoretical background}

\subsection{\textit{A contrario} decision criterion}
\subsubsection{General \textit{a contrario} framework}
\label{generalite_NFA}
\addSylv{W}e base our approach on the \textit{a contrario} formulation used by~\cite{desolneux2007gestalt, ipol.2019.263, vidal2019aggregated}; this formulation is straightforward since we work with gray-scale feature maps.
\addSylv{Assuming the naive model \addmoi{for the background} is} a centered Gaussian distribution with unit variance (hypothesis $H_0$)\addSylv{, the following} function is defined, for each tested pixel $x_i$, as: 
\begin{equation}
    f(i,x_i) = N_{test} \times \mathbb{P}_{H_0}(||X_i||^2_2 \geq ||x_i||^2_2),
    \label{eq:NFA_form}
\end{equation}
where $(X_i)_{i \in \mathbb{N}}$ is a sequence of variables \addSylv{that are assumed to} follow $H_0$,  $\mathbb{P}_{H_0}$ the associated probability \addSylv{ and $N_{test}$ is the so-called ``number of tests'' that} corresponds to the total number of analyzed observations $x_i$ (i.e. in our case the number of pixels composing the image).
As demonstrated by~\cite{grosjean2009contrario}, \addSylv{Eq.~\eqref{eq:NFA_form} defines} a Number of False Alarms (NFA) \addSylv{provided that, $ \forall \epsilon > 0,$} it is $\epsilon$-meaningful, i.e. the following condition is verified:
\begin{equation}
    \mathbb{E}[\#\{i,f(i,X_i)\leq \epsilon\}] \leq \epsilon,
\end{equation}
\addSylv{where the symbol $\mathbb{E}[.]$ stands for the mathematical expectation and $\#\{.\}$ for the cardinality of a set.} 
This property guarantees that, on average, raising a detection every time \addSylv{$f$} is \addSylv{lower than} $\epsilon$ should lead to at most $\epsilon$ false alarms. \addSylv{Thus, such a} function allows for the control of the number of false alarms. Thereafter, we call ``NFA'' the tested value $f(i,x_i)$.

\subsubsection{Multi-channel formulation}
\label{sec:nfa_formulation}
\addSylv{In~}\cite{ipol.2019.263}\addSylv{, the authors} adapted \addSylv{previous single channel formulation} to multi-channel input by \addSylv{considering} each channel independently. The obtained NFA maps are then merged together by taking the union of detections. \addSylv{In this study, we rather} reformulate the previous approach in terms of a multivariate normal distribution, as suggested by~\cite{vidal2019aggregated}. By considering a centered input $X_i$ \addSylv{with} $K$ channels, we can rewrite Eq.~\eqref{eq:NFA_form} \addSylv{ using the G}amma and upper incomplete \addSylv{G}amma functions \addSylv{(denoted $\Gamma(.)$ and $\Gamma(.,.)$ respectively)}:



\begin{equation}
    \NFA(i,x_i) = \frac{N_{test}}{\Gamma(K/2)} \times \Gamma(\frac{K}{2},\frac{1}{2} ||\Sigma^{-1/2} x_i||^2_2),
    \label{eq:NFA_form_gamma}
\end{equation}
where $\Sigma$ represents the covariance matrix of the centered variable $X_i$. Three assumptions about the feature noise are made: 1) Elliptical distribution with dependent channels: in this case, $\Sigma$ is a dense positive-definite matrix; 2) Elliptical distribution with independent channels, which leads to $\Sigma = \lambda \Delta$ where $\Delta$ is a diagonal matrix with $|\Delta|=1$ and $\lambda \in \mathbb{R}$; 3) Spherical distribution, leading to $\Sigma= \lambda I_d$ where $I_d$ is the identity matrix. In this particular case, no direction or channel is privileged in the decision process. The impact of these different forms on \addSido{training} is assessed in Appendix~\ref{ablation_SIRST}.

\subsubsection{NFA and \textit{significance}}
\label{nfa_sign_link}
\begin{figure}[ht]

  \centering
    \includegraphics[width=8cm]{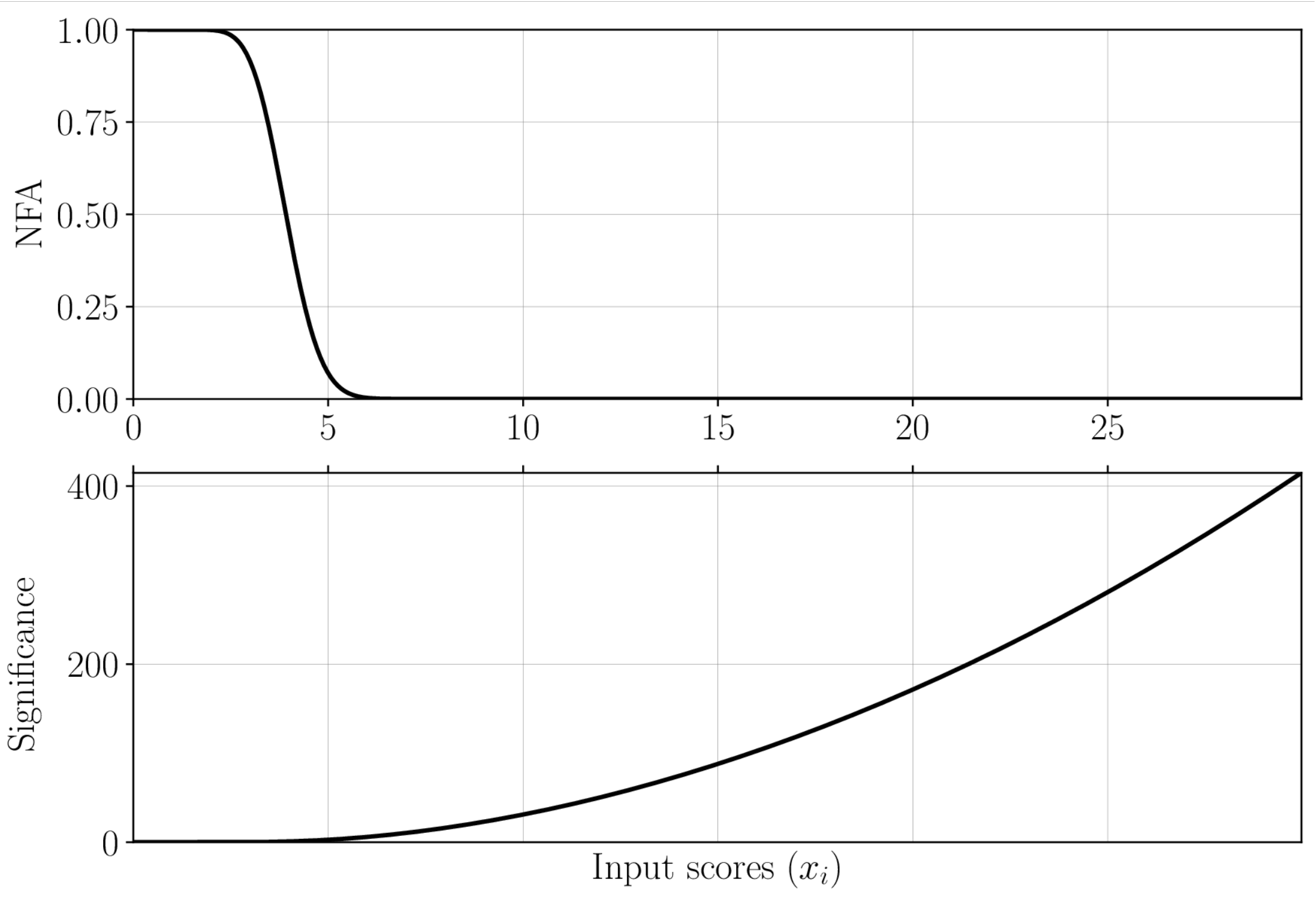}
    \caption{NFA and \textit{significance} values for a centered unit variance Gaussian variable, with $N_{test}=1$.}
    \label{fig:NFA_sign}
\end{figure}  

According to Eq.~\eqref{eq:NFA_form}, the NFA values range
from 0 to $N_{test}$. However, for large values of $x_i$, the NFA values tend towards very small values, often \addSylv{lower than} $10^{-200}$. In order to \addSylv{increase the readability of} those values, \addSylv{some authors have proposed to rather consider} the \textit{significance} $S(i,x_i)$ \addSylv{defined using} a logarithmic scaling:
\begin{equation}
    S(i,x_i) = -\ln(\NFA(i,x_i)).
    \label{eq:NFA_signif}
\end{equation}

\addSylv{Then, the \textit{significance} associated to the NFA of} Eq.~\eqref{eq:NFA_form_gamma} \addSylv{is as follows}:
\begin{equation}
    S(i,x_i) = -\ln(\frac{N_{test}}{\Gamma(K/2)}) -\ln(\Gamma(\frac{K}{2},\frac{1}{2} ||\Sigma^{-1/2}x_i||^2_2)).
    \label{eq:NFA_form_signif}
\end{equation}
NFA values and their corresponding \textit{significance} are represented on Figure \ref{fig:NFA_sign}. \addmoi{Due to rounding problems to 0, we use the approximation of the $\Gamma(a,x)$ function for $x\to +\infty$ (in practice, $x>40$) given in \cite{handbook}}:

\begin{equation}
    \Gamma(a,x) \approx x^{a-1}e^{-x}(1+\frac{a-1}{x}+\frac{(a-1)(a-2)}{x^2}).
    \label{eq:loggamma_asympt}
\end{equation}

\subsection{Attention mechanisms}

Despite the efficiency of CNNs in extracting meaningful information from an image, the translation invariance induced by convolutions seems to impair the overall understanding of the scene. 
Attention mechanisms partly circumvent this limitation by imitating the human perception and by dynamically weighting features depending on their relevance to a given final task. 
Several types of attention \addSylv{mechanisms have been proposed}, leading to a wide range of techniques discussed in~\cite{guo2022attention}. In our work, we focus on the use of channel and spatial attention mechanisms.

\subsubsection{Channel-based attention}
Channel-based attention allows \addSylv{us to} select \addSylv{the} relevant channels in \addSylv{a set of} feature maps. This concept was first presented \addSylv{in}~\cite{hu2018squeeze}, where the authors introduce a squeeze-and-excitation block made of two steps. The first one, called the squeeze step, consists in a reduction in dimension\addSylv{ality} while keeping global spatial information. Then, an excitation module allows \addSylv{for} learning channel-wise relationships, which gives rise to an attention vector that indicates the weights to \addSylv{apply} to the different channels. Several variants have been proposed to overcome SE block shortcomings. For example, \cite{eca_net} propose the Efficient Channel Attention (ECA) block, where they reduce the complexity of the fully-connected layers used in the excitation step by replacing them with a 1D convolution. \addmoi{In the following, we focus on this solution.}


\subsubsection{Spatial attention}
Unlike channel attention, spatial attention is intended to indicate \addSylv{the} regions of the image where most attention is needed. This is achieved by modeling long-range dependencies between \addSylv{the} different regions of an input. Several strategies have been proposed, including training a subnetwork to identify \addSylv{the} important regions~\cite{mnih2014recurrent, jaderberg2015spatial, dai2017deformable}, or increasing the receptive field of CNNs. \addSylv{The m}ethods based on the latest strategy use self-attention mechanisms, which were introduced in computer vision tasks by~\cite{wang2018non}. 
They lead to impressive results compared to \addSylv{the performance} achieved so far using CNNs, especially when it comes to the use of Vision Transformers (ViT) for various visual tasks~\cite{vaswani2017attention, xie2021segformer, chen2021transunet}. However, this process is computationally very expensive, and it also requires a lot of training data. \addmoi{In addition, for small object detection, the spatial dependencies are mainly local.}
In this work, we rather consider the use of local self-attention layers, and more specifically the stand-alone self-attention layers proposed by~\cite{ramachandran2019stand}. 



\section{Deep \textit{a contrario} framework}
In this section, we present our method for integrating an \textit{a contrario} decision criterion into a one-class semantic segmentation NN. Two key ingredients are needed for such module: an \textit{a contrario} block, called NFA block, and a specific activation function that allows the NN to learn from the obtained \textit{significance} scores. 
\begin{figure*}[ht]

  \centering
    \includegraphics[width=18cm]{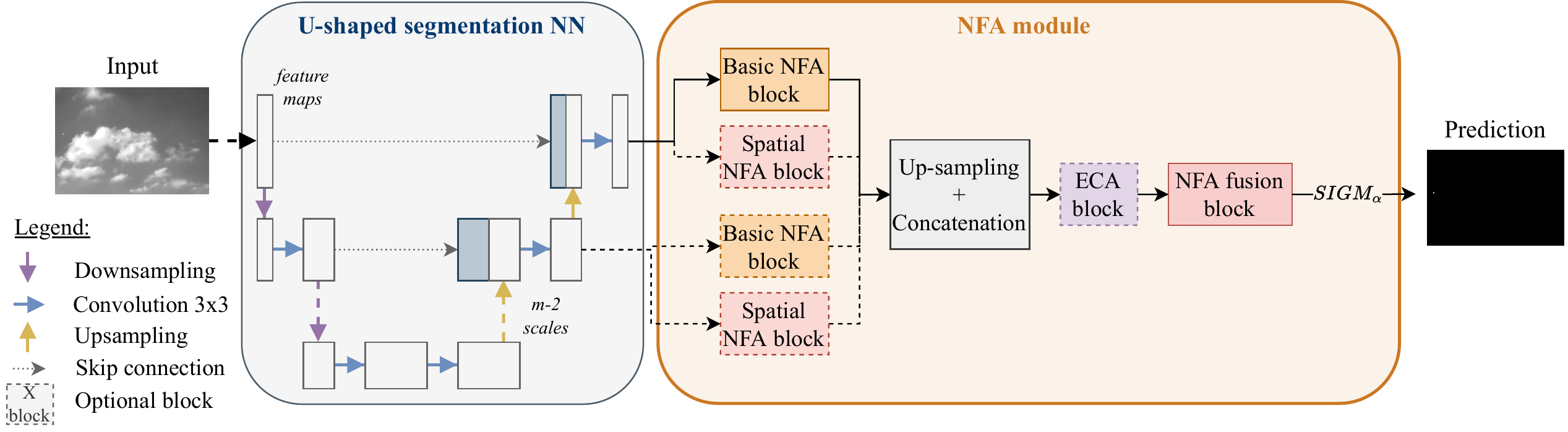}
    \caption{Diagram showing the integration of our NFA module into a U-shaped segmentation NN. Optional blocks are drawn in dotted lines. Details for ECA block can be found in the original paper \cite{eca_net}.}
    \label{fig:archi_globale}
\end{figure*}  

     \begin{figure}[ht]
         \centering
         \includegraphics[width=8.5cm]{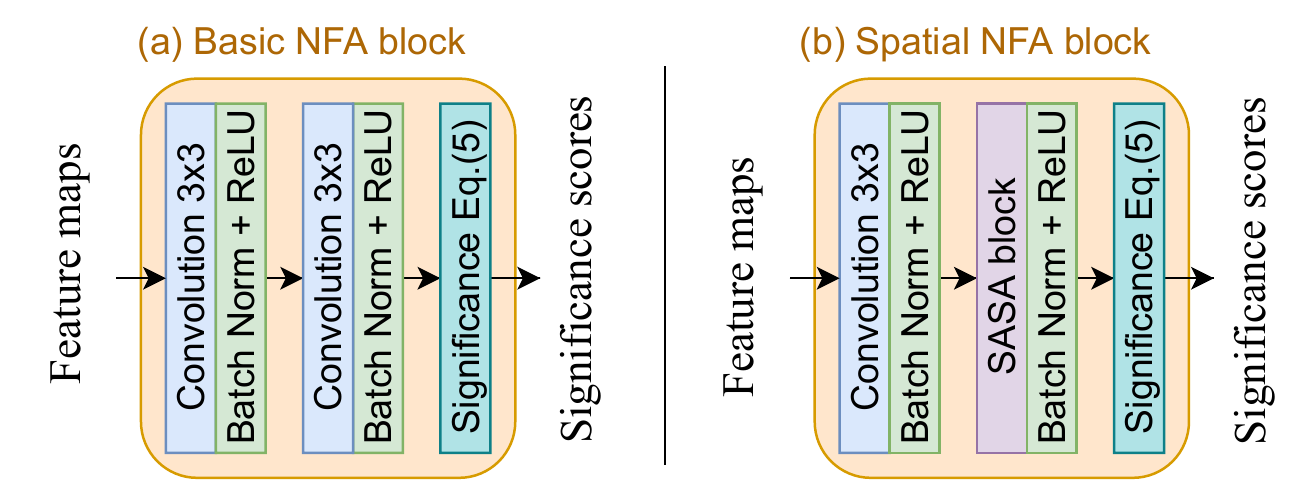}
         \caption{Diagram of (a) the basic NFA block and (b) the spatial NFA block. The details of the stand-alone self-attention (SASA) block can be found in~\cite{ramachandran2019stand}.}
         \label{fig:spatial_NFA_block}
     \end{figure}

\subsection{NFA blocks}
\subsubsection{Basic NFA block}
\label{basic_NFA}
We propose a basic NFA block that transforms \addSylv{the} feature maps into a one-channel score map \addSylv{representing \textit{significance}} \addSylv{defined by}~\eqref{eq:NFA_form_signif}. This block is described by Figure~\ref{fig:spatial_NFA_block}a.
Two convolution blocks (i.e.\addSylv{,} 2D convolution with kernel 3 followed by batch normalization and ReLU activation) are applied on the input features in order to extract \addSylv{some} relevant features for computing the NFA. The \textit{significance} scores are then computed using Eq.~\eqref{eq:NFA_form_signif}. Note that this equation is derivable, allowing the backpropagation step in the \addSylv{NN}. This NFA block can replace the segmentation head of a one-class segmentation NN. Its integration on a U-shaped NN is presented in Figure~\ref{fig:archi_globale}.

\subsubsection{Multi-scale fusion of \textit{significance} maps}
\label{multi-scale}
Many popular segmentation networks rely on encoder-decoder models \addmoi{introduced in} \cite{simonyan2014very, ronneberger2015u}. The advantage of using U-shaped NN is that we can easily extract low-level semantic feature maps and use the large-scale spatial information they contain for detecting \addSylv{some} objects of different sizes. To do so, we integrate our basic NFA block at each intermediate scale of any U-shaped NN, as illustrated in Figure \ref{fig:archi_globale}. \addSylv{Considering an original NN with $m$ scales, w}e thus obtain $m$ \textit{significance} score maps. The low-level \textit{significance} score maps are then upsampled using bilinear interpolation in order to match the NN input size. All \textit{significance} maps $ S_1, ...,S_m$ are merged together through the NFA fusion block by taking the union of all detections, as in \cite{ipol.2019.263}. This leads to the final \textit{significance} score map $S_{\mbox{\scriptsize \textit{final}}}$, defined for each pixel $i$ as follows: 

\begin{equation}
    S_{\mbox{\scriptsize \textit{final}}}(i) = \min \{S_1(i), ...,S_m(i)\}.
    \label{eq:NFA_merged}
\end{equation}


However, \addSylv{with such a} multi-scaling strategy, \addSylv{the} detections from lower and higher resolution scales have the same weight in the final \textit{significance} score map, which may increase the false alarm rate for applications where coarse scales are less relevant. We \addSylv{thus} propose to dynamically weight the impact of the different scales by learning weighting coefficients using a channel attention module. The integration of an ECA block~\cite{eca_net} before merging the \textit{significance} maps is illustrated on Figure~\ref{fig:archi_globale}.

\subsubsection{Spatial NFA block}

The basic NFA block defined in Section~\ref{basic_NFA} is designed to improve the detection of small objects. However, for small objects that are not point-shaped and are small only in one dimensionality and significantly large in other dimension(s) (e.g. cracks), geometric information is a particularly discriminating feature.
To improve performance on such objects, we design a second version of our NFA block that includes spatial attention mechanisms. Figure~\ref{fig:spatial_NFA_block}b shows \addSylv{this} block, where the second convolution layer is replaced by a stand-alone self-attention (SASA) layer~\cite{ramachandran2019stand}. \addSylv{As shown in Figure~\ref{fig:archi_globale},} \addmoi{if we add a spatial NFA block, it is done in addition to a basic NFA block. }



\subsection{NFA-friendly activation function}
\label{act_fct}
\begin{figure}[ht]

  \centering
    \includegraphics[width=8cm]{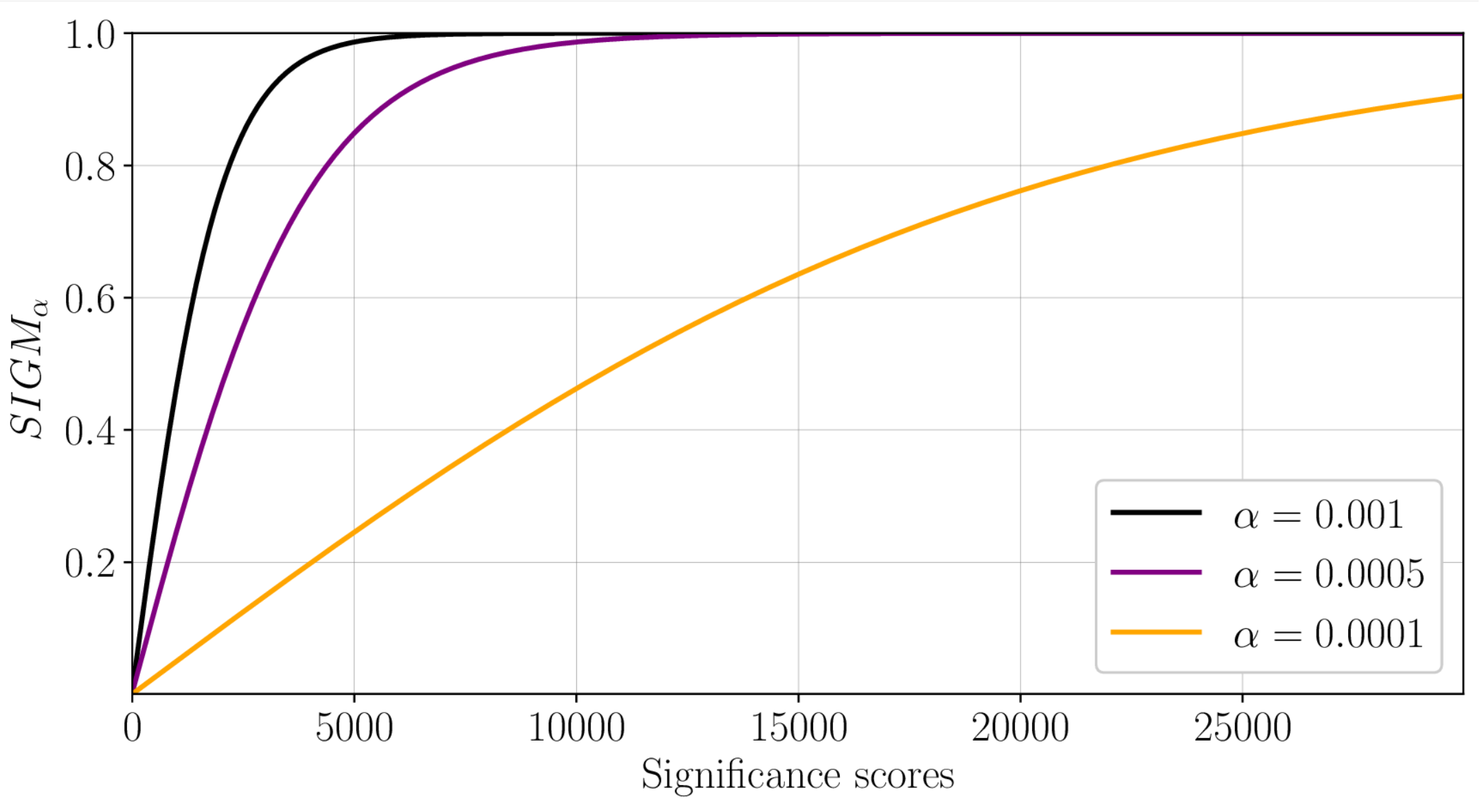}
    \caption{Variations of $\Sigm$ function defined in \eqref{eq:sigm_a}, with different values of $\alpha$. For simplicity, we choose $N_{test}=1$.}
    \label{fig:sigm_fct}
\end{figure}  

The NFA block output \addSylv{is} a \textit{significance} score map whose distribution of scores is not only asymmetric between positive and negative values, but also has a much wider dynamic \addmoi{than conventional NN output}, as explained in Section~\ref{nfa_sign_link}. Consequently, \addSylv{the} conventional symmetric activation functions, such as sigmoid function, are not suitable \addSylv{and we rather consider} the following activation function:
\begin{equation}
    \Sigm(x, N_{test}) = \frac{2}{1+e^{-\alpha (x+\ln(N_{test}))}}-1,
    \label{eq:sigm_a}
\end{equation}
where $\alpha \in$ \addSido{$\mathbb{R}$} is a parameter that allows \addSylv{us} to control the slope of the sigmoid and $N_{test}$ is the number of tests presented in \addSylv{Section}~\ref{generalite_NFA}. We represent the variations of $\Sigm$ function on Figure~\ref{fig:sigm_fct} for different values of $\alpha$. 
\addSylv{T}he higher the value of the parameter $\alpha$, the more the dynamic of the \textit{significance} scores will be \addmoi{non linearly compressed}. The sensitivity of the NN training to this parameter is studied in Appendix~\ref{ablation_SIRST}. This activation function is applied after \addSylv{having combined} all \addSylv{the} \textit{significance} maps obtained from the NFA blocks \addSylv{ computed at different scales}, as \addSylv{shown} on Figure~\ref{fig:archi_globale}. The final output scores therefore range between $0$ and $1$, which allows the \addSido{user} to apply any cost function that is suitable for binary segmentation tasks. 

Nevertheless, substituting the conventional segmentation head in a segmentation NN by the NFA module modifies the dynamic of the output scores in a way that makes the threshold usually used to binarize the segmentation map (namely, $0.5$) no longer suitable. 
We have derived the threshold on the \textit{significance} scores after observing that object pixel scores are usually greater than a few hundred while background ones are lower than $30$, which correspond to a NFA around $10^{-200}$ and thus a \textit{significance} map threshold around $500$. 
This allows us to roughly estimate an interval of threshold values to be applied on the NN predictions, that is $[0,\Sigm(500)]$. Depending on the parameter $\alpha$, this interval will be more or less wide: for example, $\alpha=0.0005$ leads to an interval of $[0,0.13]$. This shows that the threshold needs to be small enough, and a more refined choice can be achieved using a validation dataset.\\

In next sections, we integrate our NFA module on different backbones and evaluate its benefits on two applications, namely small target detection and road crack detection.

\section{Application to small target detection}
\label{sec:target_det}
\addmoi{We first evaluate the contribution of our NFA module in the case of small \addSido{infrared} target detection. This application constitutes an ideal framework for the detection of small objects: the targets have a surface area of only a few pixels, are not very contrasted compared to the background and do not present a specific structure. Most of the methods proposed to tackle this problem use semantic segmentation NN~\cite{kim_small_2018,dacremont_cnn-based_2019} rather than off-the-shelf detection NN~\cite{redmon2016you}. SOTA NN for small target detection rely on U-shaped architectures and include spatial attention mechanisms~\cite{liu_infrared_2021,li2022dense,zhang2022isnet}.}

\subsection{Assessed methods}
\label{sec:meth_sirst}
We propose to integrate our NFA module into one of the U-shaped SOTA method. We select the recent DNANet \cite{li2022dense}, which has shown impressive performance on \addmoi{widely used small target detection datasets}. DNANet is composed of two parts: a dense-nested U-shaped backbone (DNSC), that allows the feature extraction step, and a feature pyramid fusion module (FPFM), which allows a multi-scale fusion of intermediate outputs from the backbone. \addmoi{We substitue the FPFM block by our NFA module, and we evaluate its contribution with respect to the backbone DNSC (ResNet-18 version) and DNANet.} We also extend our experiments to the use of a classical backbone, namely ResUnet \cite{zhang2018road}, to show the generalization of our method to another backbone that is not specifically designed for small target detection.

\addSylv{For our NFA module, w}e set the $\alpha$ parameter in Eq.\eqref{eq:sigm_a} to $0.0005$, \addmoi{as it has shown to lead to best results in Appendix~\ref{ablation_SIRST}}. Based on the discussion in Section~\ref{act_fct} and results obtained on the validation dataset, we fix the map binarization threshold to $0.1$.
For the \addmoi{baselines}, the detection threshold is set to $0.5$ as suggested in the original paper. All networks are trained from scratch \footnote{\addmoi{We used the official PyTorch~\cite{NEURIPS2019_9015} implementation of DNANet \href{https://github.com/YeRen123455/Infrared-Small-Target-Detection}{https://github.com/YeRen123455/Infrared-Small-Target-Detection}}} on Nvidia RTX6000 GPU for $1000$ epochs using the Soft-IoU loss function \cite{rahman2016optimizing}. The latter is optimized by Adagrad method with the Cosine Annealing scheduler, using the same parameters as in~\cite{li2022dense}.
The learning rate is set to $0.05$ for DNSC and DNANet as suggested in the original paper. For DNSC+NFA, we found that decreasing the learning rate allows for better convergence; we thus fix it to $0.03$. 

\subsection{Dataset and evaluation metrics}
\label{setup}
 We conduct our experiments on NUAA-SIRST dataset \cite{dai2021asymmetric}, which is \addSylv{one of the few infrared small target datasets publicly \addSido{released} and} widely used in the literature. 
 This dataset contains 427 infrared images, and we split it into training, validation and tests sets \addmoi{using }a ratio of 60:20:20. We use the same \addmoi{pre-processing steps as those} proposed in~\cite{li2022dense}.

For the evaluation, we mainly focus on object-level metrics as suggested by~\cite{li2022dense}. 
We therefore compute the Precision (Prec.), Recall (Rec.) and F1 score (F1) at object-scale. We give the area under the object-level Precision-Recall curve, namely Average Precision (AP), which allows \addSylv{us} to \addSylv{free from the detection} threshold. We also compute the number of false alarms \addSylv{(still at object-level)} per image (FA/image). \addSylv{From the predicted binary segmentation map, targets are individually labeled using a 8-connectivity connected component module.} \addmoi{A detected object is counted as a true positive (TP) if it has an IoU of at least 5\% with the ground truth. This low-constrained condition is due to the fact that a small shift in the number of predicted pixels leads to a large deviation in the IoU while not being a relevant indicator for object-level evaluation.} 


\addSylv{In the tables, the presented} results \addSylv{have been} averaged over \addSylv{three} distinct training sessions and they are given in the form $\mu \pm \sigma$, where $\mu$ is the mean and $\sigma$ the standard deviation.

\subsection{Results}

\subsubsection{NFA module improves the precision}

\begin{table*}[t] 
\centering
 \caption{\addmoi{Object-level F1, AP, Prec., Rec., and FA/image achieved by the compared methods }\addSido{on NUAA-SIRST.} \addmoi{Best results are in bold and second best results are underlined.}}

  \begin{tabular}{|c|c|c||c|c|c|c|} 
  \hline

    \textbf{Method} & \textbf{F1 (\%)}  & \textbf{AP (\%)} & \textbf{Prec. (\%)} & \textbf{Rec. (\%)}   & \textbf{FA/image} &\textbf{\#Params (M)}\\
  
   \hline
   \textbf{DNSC } & $96.2 \pm 1.4$ & $95.6 \pm 1.1$  &  $95.0 \pm 2.8$ & $\textbf{97.6} \pm 0.6$  &  $0.06 \pm 0.03$ &   4.67\\
   \hline
   \textbf{DNSC + FPFM (DNANet)} & $ \underline{96.9} \pm 0.5 $  & $\underline{98.1} \pm 1.2$  & $\underline{96.6} \pm 1.5$ & $97.2 \pm 0.6$  & $\underline{0.04} \pm 0.02$ & 4.70\\
   \hline
   \textbf{DNSC + NFA (Ours)}  & $\textbf{97.6} \pm 0.3$   &  $\textbf{98.4} \pm 0.7$ & $\textbf{97.8} \pm 0.0$ & $\underline{97.5} \pm 0.6 $& $\textbf{0.02} \pm 0.00$ &  4.76\\
   \hline
  \end{tabular}
  
  \label{res_SIRST}
\end{table*}
\begin{figure}[ht]
         \centering
         \includegraphics[width=8.5cm]{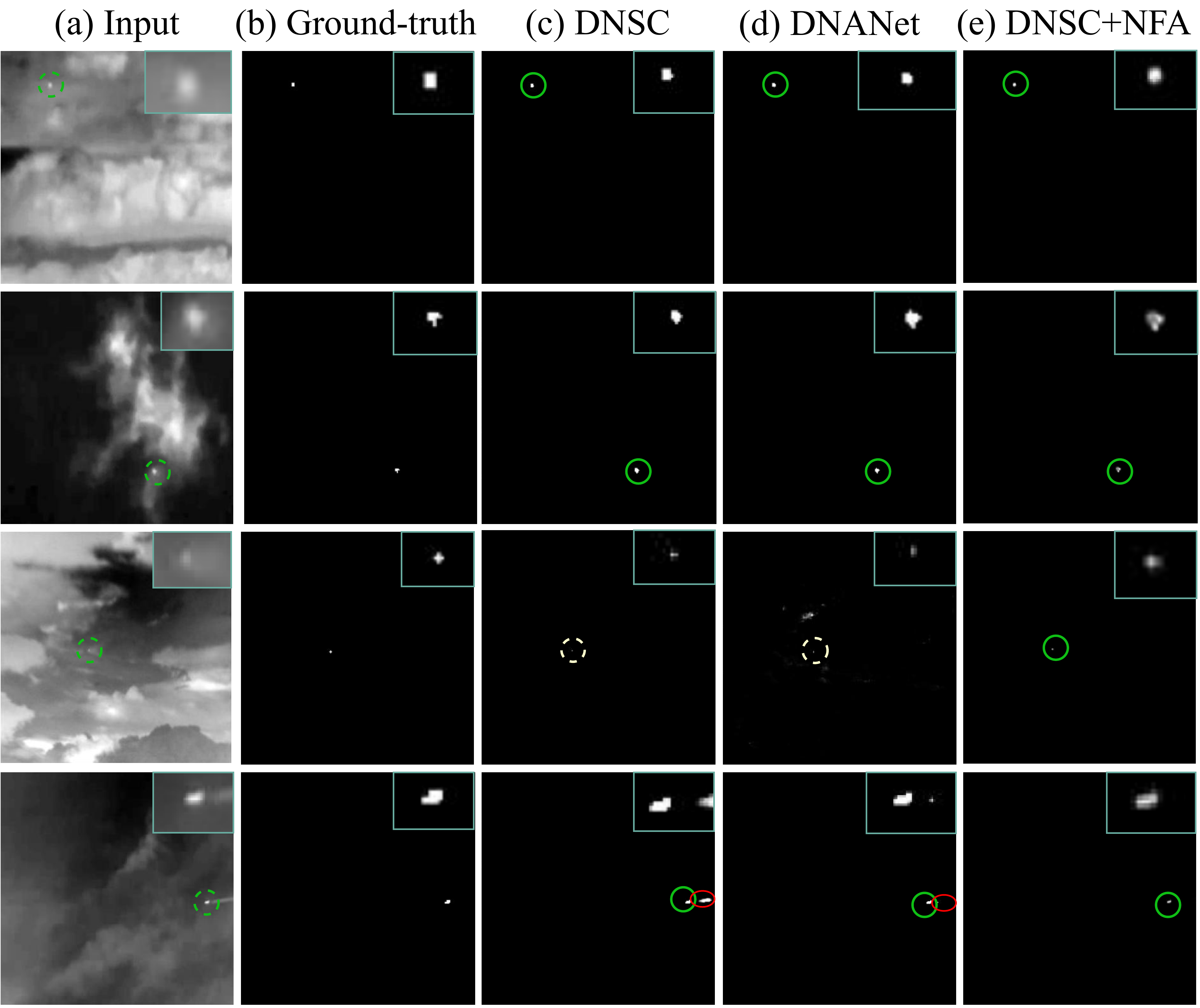}
         \caption{Qualitative results obtained with different detection methods (columns (c) to (e)) on NUAA-SIRST dataset. Good detections and false positives are circled in red and green lines respectively, and missed detections in light yellow dotted lines. The ground truth is circled in green dotted lines \addSido{in column (a)}. A zoom on the targets is displayed in the top right corner.}
         \label{fig:sirst_viz}
     \end{figure}

Table~\ref{res_SIRST} \addSylv{shows} the performance for the three compared methods. DNSC+NFA clearly improves the baseline global performance, and even outperforms DNANet. 
More specifically, \addSylv{since} the NFA layer controls the number of false alarms\addSylv{, the Precision appears significantly improved}, while keeping the number of correctly detected targets \addSylv{(Recall criterion)} at \addSylv{almost} the same level. \addSylv{Besides, the global measure provided by Average Precision clearly underlines the robustness of the proposed network with respect to the detection threshold.}
\addmoi{Note that adding the NFA module in DNSC greatly improves the stability of the training, as evidenced by the decrease in the standard deviation in the results.}
Furthermore, the NFA layer adds less than $0.1$ million parameters to the initial model, which is negligible \addmoi{with respect to} the benefits deriving therefrom.

The results obtained for ResUnet+NFA are detailed in Appendix \ref{resunet_sirst}, and the conclusions are the same as for DNSC+NFA.

\addmoi{Figure~\ref{fig:sirst_viz} illustrates some predictions (output scores before threshold) on challenging scenes, where the contribution of the NFA module can clearly be seen. 
For example, the target of the 3rd line is particularly small and blurred in the background, which does not affect the performance of the NFA module, unlike the other methods. Moreover, the baseline methods mistakenly detect the aircraft contrail (line 4). The NFA module not only allows for better detection of small objects in particularly difficult scenes, but also provides robustness with respect to challenging environments.}

\subsubsection{Overconfidence, did you say? }
\begin{figure}[ht]
     \centering
         \includegraphics[width=8cm]{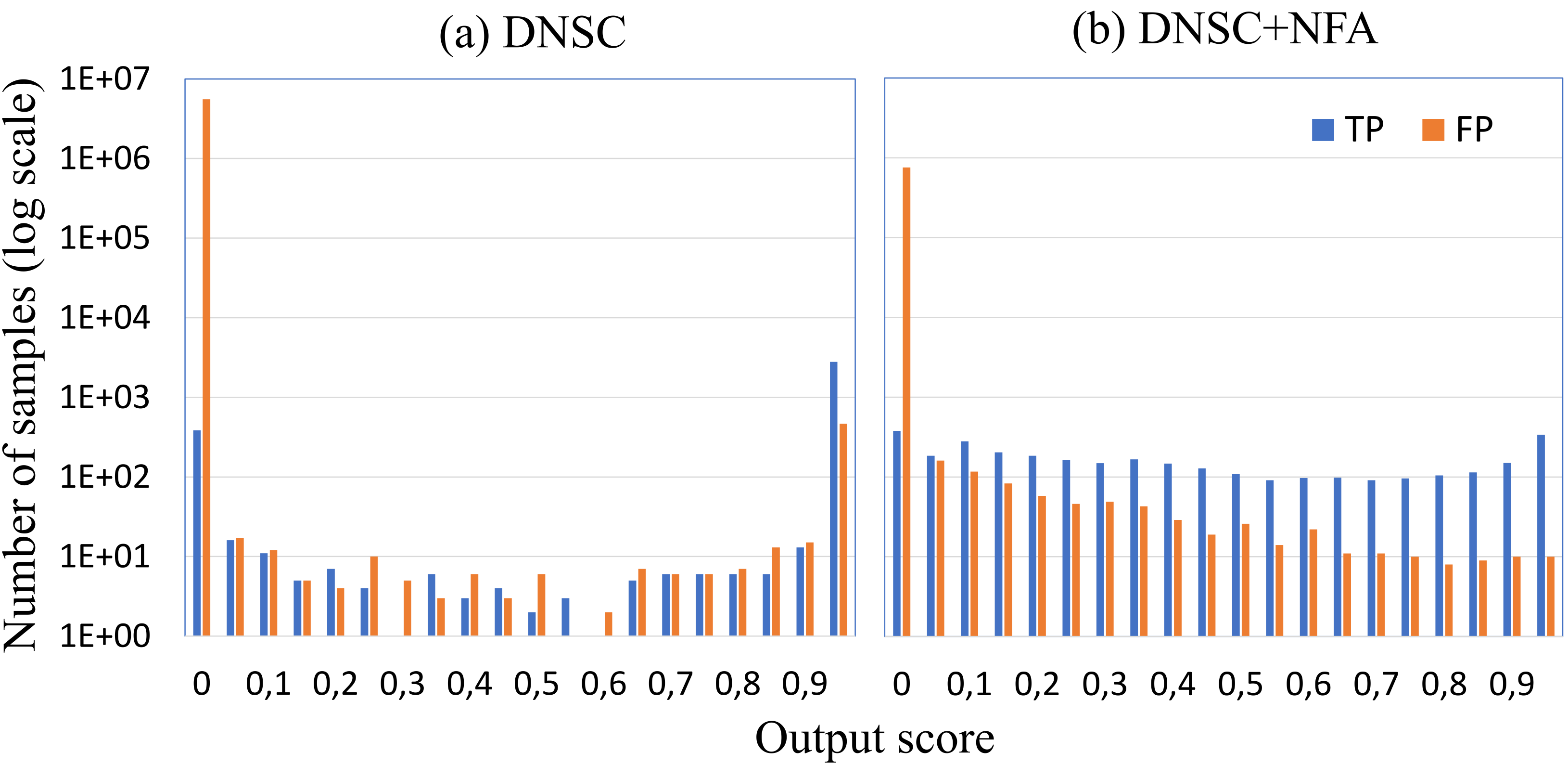}
        \caption{Output scores histograms for (a) DNSC and (b) DNSC+NFA.}
        \label{fig:histo}
\end{figure}
\begin{figure*}[t]

  \centering
    \includegraphics[width=17cm]{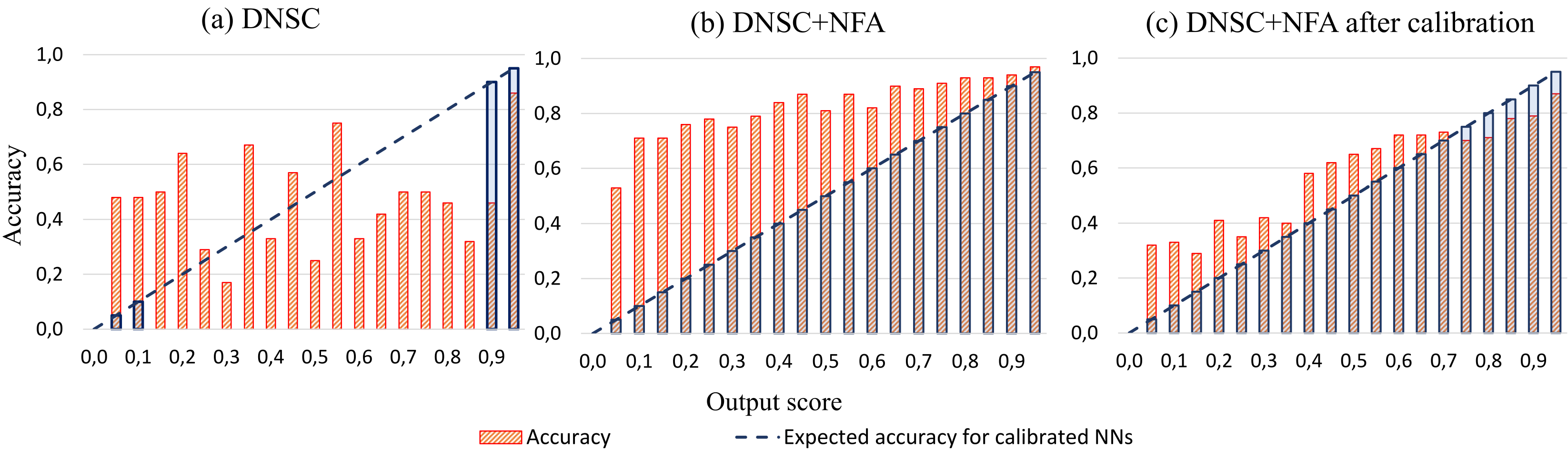}
    \caption{Variations in accuracy as a function of output scores for (a) DNSC, (b) DNSC+NFA with $\alpha=0.0005$, and for (c) DNSC+NFA after calibration (i.e. $\alpha=0.003$).}
    \label{fig:after_calibration}
\end{figure*}

\addSylv{M}ost recent neural networks tend to be overconfident \addSido{as outlined in} \cite{guo2017calibration} 
The pixel-level histogram of output scores \addSylv{shown on} Figure~\ref{fig:histo}a \addmoi{illustrates} this phenomenon for DNSC network, where all pixel values on the final score map are either very close to $0$ or to $1$. The impact of NFA layer can clearly be seen on the corresponding histogram in Figure~\ref{fig:histo}b: TP are uniformly spread all over the confidence scores and the number of false positives (FP) decreases monotonically as the score level increase\addSylv{s}. Consequently, we can see \addSylv{on Figure~\ref{fig:after_calibration}b that} the \addSylv{achieved} scores \addSylv{are much more informative} since the accuracy  increas\addSylv{es} with the score. However, even though the NFA module prevents the network from being overconfident, DNSC+NFA is not calibrated. \addmoi{This calibration can be achieved by playing with the parameter $\alpha$ from Eq.\eqref{eq:sigm_a}, although we recommend to adjust this parameter after training as it has been observed in Appendix \ref{ablation_SIRST} that keeping a wider dynamic for output scores facilitates learning. Setting $\alpha = 0.003$ leads to the a more calibrated histogram on Figure \ref{fig:after_calibration}c, while keeping an increasing accuracy. Therefore, adding the NFA module to a segmentation NN allows \addSylv{us} to \addSido{obtain a nearly calibrated network} without the need of complex methods. The output scores are also relevant, which is a step towards AI interpretability. }


\subsubsection{Ablation study}
An ablation study is performed to evaluate the contribution of each main component of our NFA module, which details can be found in Appendix~\ref{ablation_SIRST}. Our conclusions are as follows: 1) regarding the assumptions made about the form of the covariance matrix, $\Sigma=\lambda \Delta$ appears to be the most relevant hypothesis; 2) as far as target detection at an object-level is concerned, adding a regularization term that forces the NN to minimize the gradient of the predicted scores is essential to avoid object fragmentation; 3) taking into account the information given by low-level feature maps improves the global performance but leads to a decrease in precision; \addSido{4) this decrease can be prevented by} weighting the different scales with channel attention. It also gives an insight on which level feature maps contributes the most to the decision. 

\section{Extension to crack detection}
\begin{table*}[t] 
\centering
 \caption{Comparison of ResUnet and \addSido{ResUnet + }NFA on crack detection. Metrics are computed at pixel-level.}

  \begin{tabular}{|c|c|c||c|c|c|} 
  \hline
  
\textbf{Method}  & \textbf{F1 (\%)} & \textbf{AP (\%)} & \textbf{Prec. (\%)} & \textbf{Rec. (\%)}\\
    
   \hline
   
   \textbf{ResUNet} &  $85.6 \pm 0.4 $  & $85.2 \pm 0.2 $   & $\textbf{94.1} \pm 0.6 $   & $78.5 \pm 0.9 $   \\
   \hline
  
   \textbf{ResUNet + NFA } &  $\textbf{87.4} \pm 0.1 $  & $\textbf{96.7} \pm 0.2 $ & $93.8 \pm 0.6 $   & $\textbf{81.9 }\pm 0.5 $    \\
   \hline
  \end{tabular}
  
  \label{cracks_res}
\end{table*}
\addmoi{
We have shown in previous section that the NFA module can improve the performance of a segmentation NN specifically designed for small target detection. This allowed us to obtain state-of-the-art results on an application that represents an ideal framework for small object detection. Now, we propose to \addSylv{expand the boundaries of previous framework}. For this purpose, we integrate our method on a classical semantic segmentation backbone and we apply it to road crack detection. \addSylv{T}he latter application is still a small object detection problem \addSylv{ since} the cracks are very thin and \addSylv{their pixels are very few with respect} to the background class. Crack detection is challenging because of the textured background and road artifacts that lead to numerous false alarms. Some generic deep learning approaches have been tested on this application, and are mainly based on classical segmentation NN~\cite{konig2019convolutional,fan2020automatic,li2021pavement,Rill-GarciaDD22}.}

\subsection{Assessed methods}

We take as a baseline a classical segmentation backbone, namely a Unet with a ResNet encoder (ResUnet, \cite{zhang2018road}). \addSylv{Note that, f}or crack detection, geometric information is  crucial \addSylv{since the} cracks \addSylv{exhibit a specific} shape. Therefore, \addSylv{we take the opportunity given by crack detection to} evaluate the contribution of the spatial NFA block in the NFA module. \addmoi{Based on the ablation study conducted in Appendix~\ref{ablation_SIRST}, we use the multi-scale NFA module with $\Sigma=\lambda \Delta$ (Eq.~\eqref{eq:NFA_form_gamma}) and set the parameter $\alpha$ in Eq.~\eqref{eq:sigm_a} to $0.0005$.} A test on the validation dataset allows us to set the detection threshold to $0$ for ResUnet+NFA. For a fair comparison with the baseline, whose optimal threshold no longer seems to be $0.5$, we choose the threshold for the baseline \addSylv{based on} the validation dataset. Both methods are trained for $700$ epochs using the same loss and optimizer as in Section~\ref{setup}. ResUnet is trained with a learning rate of $0.01$, and we lower the learning rate for ResUnet+NFA to $0.005$.


\subsection{Dataset and evaluation metrics}


We train and evaluate all methods on Crack Tree dataset \addSido{from} \cite{zou2012cracktree}. It is composed of $206$ real pavement images, and it includes various types of cracks. 
Because very few data is available, \addSylv{the} algorithms are trained using $120$ images only. This frugal setting adds \addSylv{some} challenge to the application. Finally, $36$ images are used for the validation step, and $50$ for testing. \addSylv{The m}ethods are evaluated using pixel-level metrics, namely precision, recall, F1 score and average precision. However, as stated in~\cite{zou2012cracktree}, the annotations do not accurately report crack thickness. Therefore, \addSylv{like} in the original paper, we adopt a tolerance margin of 2 pixels, and we ignore pixels predicted within this margin.


\subsection{Results}

\subsubsection{NFA module leads to better performance}


\begin{figure}[ht]
         \centering
         \includegraphics[width=8.9cm]{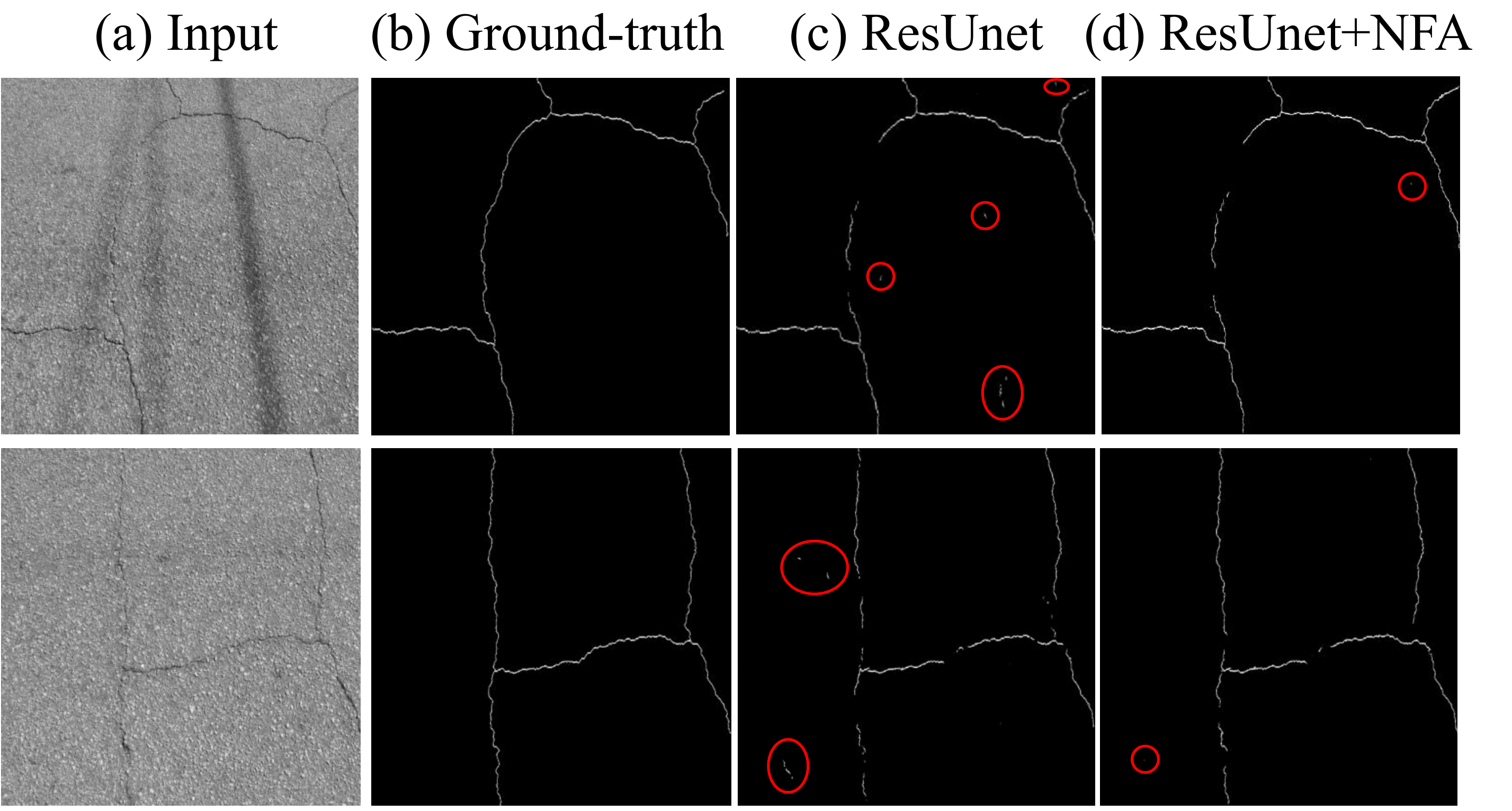}
         \caption{Qualitative results obtained with different detection methods on Crack Tree dataset. False positives are circled in red.}
         \label{fig:cracks_viz}
     \end{figure}
Table~\ref{cracks_res} shows the performance of the evaluated methods on \addSylv{Crack Tree dataset}, at a pixel-level. It is clear that the NFA module contributes in improving the baseline. Indeed, the F1 score is almost $2\%$ higher when using ResUnet+NFA than the baseline. More precisely, we \addSylv{observe} a very clear improvement in the average precision, which confirms the ability of the NFA module to control the number of false alarms even at a pixel level. We notice that the recall is also \addmoi{significantly} improved.
Figure~\ref{fig:cracks_viz} \addmoi{shows \addSylv{some} results obtained by the different methods on \addSylv{two different} crack \addSylv{examples}. \addSylv{T}he NFA module \addSylv{appear}s more robust to the presence of shadows or textures on the road, \addSylv{since} less false alarms \addSylv{ are observed}. }

\subsubsection{Contribution of attention mechanisms}

We \addSylv{have} evaluate\addSylv{d} the contribution of the different attention mechanisms\addSylv{, spatial one and channel attention,} on \addSylv{crack detection}. The results are detailed in Appendix~\ref{ablation_crack}. \addSylv{They show} that \addSylv{the} channel attention is still necessary to balance the weights of the different scales, and\addSylv{, based on observed weight values,} that low-level features are even more important for this application. Moreover, spatial attention is, as expected, of primary importance to prevent from irrelevant false positives. Finally, combining both channel and spatial attention mechanisms leads to even better and more \addmoi{robust} results for crack detection.

\section{Conclusion}
In this paper, we propose an add-on NFA module to improve small object detection through semantic segmentation NN. In addition to enhancing the feature map responses of small objects as proposed in the literature, our module allows for the control of the number of false alarms by exploiting the unexpectedness of small objects. 
We have experimentally demonstrated the superiority
of our method over 
state of the art methods in the case of small target detection, as well as its generalization to the detection of thin objects \addmoi{via} the application \addSylv{to} crack detection. Our NFA module also provides interpretable results, which is essential in many real-world applications.

\newpage
\bibliography{example_paper}
\bibliographystyle{icml2023}

\newpage
\appendix
\onecolumn

\section{Results of ResUnet and ResUnet+NFA on small target detection}
\label{resunet_sirst}
\begin{table*}[ht] 
\centering
 \caption{Comparison of ResUnet and \addSido{ResUnet + }NFA on small target detection. Metrics are computed at object-level.}

  \begin{tabular}{|c|c|c||c|c|c|} 
  \hline
  
\textbf{Method}  &\textbf{F1 (\%)}  & \textbf{AP (\%)} & \textbf{Prec. (\%)} & \textbf{Rec. (\%)}   & \textbf{FA/image}\\
    
   \hline
   
   \textbf{ResUNet} &  $93.2 \pm 0.9 $  & $90.3 \pm 2.4 $   & $90.6 \pm 2.1 $   & $\textbf{96.3} \pm 0.6 $ &  $0.11 \pm 0.03 $ \\
   \hline
  
   \textbf{ResUNet + NFA } &  $\textbf{95.4} \pm 1.3 $  & $\textbf{96.1} \pm 1.9 $ & $\textbf{94.6} \pm 2.2 $   & $96.2\pm 0.6 $ & $\textbf{0.06} \pm 0.03 $    \\
   \hline
  \end{tabular}
  
  \label{SIRST_resunet}
\end{table*}


\addmoi{In this section, we extend the experiments conducted in Section~\ref{sec:target_det} to another conventional NN, namely ResUnet. We evaluate the benefits of our NFA module for a NN that is not specifically designed for small target detection.}

Table~\ref{SIRST_resunet} presents the results obtained when adding the NFA module to a conventional NN for small target detection. As observed in Section~\ref{sec:target_det}, the NFA module greatly improves global performance (the F1 score is more than $2\%$ higher), which is mainly explained by an increase in average precision. Results obtained when adding the NFA module on a conventional segmentation NN are only few percents lower from what can be obtained by a SOTA segmentation NN specifically designed for small target detection (see Table~\ref{res_SIRST} for comparison). This shows that, although careful design of the feature extractor is essential to improve performance, the choice of decision criterion is \addSylv{also very} important, especially in the case of small object detection.

\section{Ablation and sensitivity studies on small target detection}
\label{ablation_SIRST}
\begin{table}[ht] 
\centering
 \caption{Ablation study performed on NUAA-SIRST (object-level metrics). We evaluated the different forms of the covariance matrix $\Sigma$ and compared the benefits of multi-scaling, adding a regularization term and using channel attention in our NFA module.}

  \begin{tabular}{|c|c|c|c|c|c|} 
  \hline

     \textbf{$\Sigma$ (Eq.\eqref{eq:NFA_form_gamma})} & \textbf{MS}& \textbf{ECA}&\textbf{Reg.}& \textbf{F1 (\%)}  & \textbf{AP (\%)} \\
    \hline
     
     \textbf{$\Sigma=\lambda I_d$} &  & &\checkmark&$96.0 \pm 0.9$ & $97.6 \pm 0.9$ \\

      \hline
     \textbf{Dense $\Sigma$} &  & &\checkmark&$95.1 \pm 1.3$ & $96.3 \pm 1.0$ \\
  
   \hline
   \textbf{$\Sigma=\lambda\Delta$} &  & &\checkmark& $96.9 \pm 0.5$ &  $\underline{98.6} \pm 1.0$  \\
   \hline

   \textbf{$\Sigma=\lambda\Delta$} & \checkmark&  & & $\underline{97.2} \pm 0.6 $  & $95.6 \pm 2.3 $  \\ 
    
   \hline
  \textbf{$\Sigma=\lambda\Delta$} & \checkmark&  & \checkmark& $\underline{97.2} \pm 0.6 $  & $97.2 \pm 0.6 $ \\ 
    
   \hline

  \textbf{$\Sigma=\lambda\Delta$} & \checkmark&\checkmark &\checkmark& $\textbf{97.6} \pm 0.3 $ &  $\textbf{99.0} \pm 0.8 $ \\
    
   \hline
  \end{tabular}
  
  \label{tab:ablation}
\end{table}

Tables~\ref{tab:ablation} and~\ref{tab:sensitivity} present the ablation and sensitivity studies performed on small target detection. The conclusions are summarized in the five following points.

\textbf{a) Assumptions made on the covariance distribution - }
\addSylv{In Section~\ref{sec:nfa_formulation}, we present three different forms for the covariance distribution $\Sigma$ in Eq.\eqref{eq:NFA_form_gamma}, corresponding to three different assumptions about feature noise: spherical distribution, elliptical distribution with independent channels or components, elliptical distribution with dependen\addmoi{t} channels. }
The first two lines of Table~\ref{tab:ablation} show that assuming a \addSylv{spherical} distribution \addSylv{assumption} leads to worse results\addSylv{, as does the channel-dependence assumption}. 
\addSylv{To explain this, one has to remind} that in deep learning, in order to disentangle causal factors, a series of filters are applied 
to extract \addSylv{the} relevant characteristics. Each filter  extract\addSylv{s} a particular feature \addSylv{represented by} a channel in the following feature maps. Depending on the downstream task, some features will be more or less relevant. The relevant information is therefore not equally distributed over all the channels of the feature maps.
\addSylv{Besides, estimat\addmoi{ing} the full covariance matrix in high dimensionality may be numerically unstable, while the correlation between extracted features remains low. This can explain why the independent elliptical distribution appears as the more relevant hypothesis.}

\textbf{b) Adding a regularization term prevents from object fragmentation - }
\addSylv{According to Table~\ref{tab:ablation}, regularization improves AP criterion. Indeed, without it, }the objects predicted by \addSido{DNSC}+NFA do not have homogeneous pixel values, and an increase in the threshold leads to fragmentation of the detected objects\addSylv{, making the results sensitive to this threshold and degrading the AP}. 
Adding a regularization term that forces the network to minimize the gradient of the predicted scores prevents this issue. 

\textbf{c) Importance of multiscaling - }
Adding information from low-level features helps the network in detecting objects of various size. This is the case for the baseline DNSC, \addSylv{whose} multiscale version is DNANet, and it has been confirmed when adding NFA layers to the 5 scales of DNSC \addSylv{and considering the F1 criterion} in Table~\ref{tab:ablation}. 
However\addSylv{, F1 increase is at the cost of a decrease of the AP criterion since introducing}
low-scale features may bring out more false positives for lower thresholds.

\textbf{d) Channel attention \addSylv{allows} for a better balancing of the scales - }
To tackle \addSylv{previous} issue, we introduced a channel attention layer before merging the different scales, that is, ECA block. Table~\ref{tab:ablation} clearly shows the superiority of \addSylv{the} NFA \addSylv{module} when adding this step. It \addmoi{noticeably} improves the average precision \addSylv{ as well as the F1 score, by} reduc\addSylv{ing} the object false alarm rate.

\addSylv{L}ooking at multiplying factors computed by channel attention layer\addSylv{, we observe that, f}or the application of small target detection,  the high-level features are of primary importance\addSylv{: their weight is} about $0.99$ \addSylv{when t}he weight of lower-level \addmoi{feature maps is about } $20$ \addSylv{ times less}, though they still contribute to the decision. 

\begin{table}[ht] 
\centering
 \caption{Sensitivity study made on the parameter $\alpha$ (Eq.\ref{eq:sigm_a}). Object-level metrics are given.}

  \begin{tabular}{|c|c|c|c|} 
  \hline

    $\alpha$ &\textbf{F1 (\%)}  & \textbf{AP (\%)}  \\
     \hline
     $0.0001$ & $96.4 \pm 1.4$ &  $95.1 \pm 0.1$\\
   \hline
    $0.0005$ &  $\textbf{97.4} \pm 0.6$  & $\textbf{97.2} \pm 0.6$ \\
   \hline
   $0.001$ & $96.7 \pm 0.2 $ & $94.8 \pm 1.0$ \\
   \hline
    
   \hline
  \end{tabular}
  
  \label{tab:sensitivity}
\end{table}
\textbf{e) $\alpha$ \addSylv{default value} - }
As explained in Section~\ref{act_fct}, the choice of $\alpha$ has an impact on the range of optimal thresholds for score map binarization. We tested \addSylv{three} values of $\alpha$, which moves the upper bound for thresholds from 0.02 to 0.3. \addSylv{According to Table~\ref{tab:sensitivity},} $\alpha=0.0005$ leads to be\addSylv{st} performance, and we recommend to use this value. 

\section{Ablation study on crack detection}
\label{ablation_crack}
\begin{table}[ht] 
\centering
 \caption{Ablation study performed on Crack Tree dataset (pixel-level metrics).}

  \begin{tabular}{|c|c|c|c|} 
  \hline

     \textbf{ECA}& \textbf{SASA}& \textbf{F1 (\%)} & \textbf{AP (\%)}   \\
   \hline
   
   &   &  $86.4 \pm 0.1 $  & $95.8 \pm 0.2 $\\
    \hline
   \checkmark &   & $\underline{87.2} \pm 0.3 $  & $96.4 \pm 0.1 $\\
   \hline
     &\checkmark   &   $\underline{87.2} \pm 0.3 $  & $\textbf{96.8 }\pm 0.2 $\\
   \hline
    \checkmark & \checkmark  &  $\textbf{87.4} \pm 0.1 $  & $\underline{96.7} \pm 0.2 $ \\

   \hline
  \end{tabular}
  
  \label{cracks_ablation}
\end{table}

Table~\ref{cracks_ablation} presents the ablation study performed on crack detection. We evaluate the contribution of channel attention and spatial attention mechanisms separately. Our conclusions are summarized in the following points. 

\textbf{a) Channel attention \addSylv{allows} for a better balancing of the scales - }
As observed in Appendix~\ref{ablation_SIRST} for small target detection, integrating the ECA block into our NFA module improves the results for both F1 and average precision. When looking at \addSylv{the} multiplying factors computed by \addSylv{the} channel attention layer, we notice that the high-level feature map is of primary importance in the decision process\addSylv{. However,} unlike in the \addSylv{small target detection}, \addSylv{the} deeper level feature maps almost equally contribute to the prediction (multiplying factor\addSylv{s} \addSylv{all around} $0.6$). \addSylv{Indeed, the} low resolution feature maps contain \addSylv{some} useful information to describe large objects, while \addSylv{the} high-level feature maps are meant for capturing \addSylv{the} smaller details \addSido{as outlined in} \cite{lin2017feature}.

\textbf{b) Spatial attention has a \addSylv{very} significant impact on performance - }
As expected, \addSylv{the} spatial attention block (SASA block) helps detecting  precisely large objects. Indeed, \addSylv{thanks to spatial attention,} the average precision is considerably improved\addSylv{:} the shape of the cracks \addSylv{is estimated in an accurate way} while eliminating some false positives. 

\addSylv{Finally, c}ombining both spatial and channel attention leads to even better and more stable results.


\end{document}